\documentclass[english]{svproc}
\usepackage[T1]{fontenc}
\usepackage[latin9]{inputenc}
\usepackage{color}
\usepackage{babel}
\usepackage{array}
\usepackage{longtable}
\usepackage{rotating}
\usepackage{url}
\usepackage{multirow}
\usepackage{amsbsy}
\usepackage{amstext}
\usepackage{amssymb}
\usepackage{graphicx}
\PassOptionsToPackage{normalem}{ulem}
\usepackage{ulem}

\makeatletter

\newcommand{\noun}[1]{\textsc{#1}}
\providecommand{\tabularnewline}{\\}

\usepackage{bibspacing}

\usepackage{lmodern}

\mainmatter              

\titlerunning{Robustness of GLVQ models against adversarial attacks}  
\author{Sascha Saralajew\inst{1} \and Lars Holdijk\inst{1,2} \and
Maike Rees\inst{1} \and Thomas Villmann\inst{3}}
\authorrunning{Sascha Saralajew et al.} 
\tocauthor{Sascha Saralajew, Lars Holdijk, Maike Rees and Thomas Villmann}
\institute{
Dr. Ing. h.c. F. Porsche AG, Weissach, Germany,\\
\email{<firstname>.<lastname>@porsche.de}
\and
University of Groningen, Groningen, Netherlands
\and
Saxony Institute for Computational Intelligence and Machine Learning, \\ University of Applied Sciences Mittweida, Mittweida, Germany,\\
\email{thomas.villmann@hs-mittweida.de}
}

\makeatother

\begin{document}

\title{Robustness of Generalized Learning Vector Quantization Models against
Adversarial Attacks}
\maketitle
\begin{abstract}
Adversarial attacks and the development of (deep) neural networks
robust against them are currently two widely researched topics. The
robustness of Learning Vector Quantization (LVQ) models against adversarial
attacks has however not yet been studied to the same extent. We therefore
present an extensive evaluation of three LVQ models: Generalized LVQ,
Generalized Matrix LVQ and Generalized Tangent LVQ. The evaluation
suggests that both Generalized LVQ and Generalized Tangent LVQ have
a high base robustness, on par with the current state-of-the-art in
robust neural network methods. In contrast to this, Generalized Matrix
LVQ shows a high susceptibility to adversarial attacks, scoring consistently
behind all other models. Additionally, our numerical evaluation indicates
that increasing the number of prototypes per class improves the robustness
of the models.
\end{abstract}

\section{Introduction}

The robustness against adversarial attacks of (deep) neural networks
(NNs) for classification tasks has become one of the most discussed
topics in machine learning research since it was discovered \cite{GoodfellowEtAl:ExplainingAndHarnessingAdversarialExamples:arXiv2015,Szegedy2013}.
By making almost imperceptible changes to the input of a NN, attackers
are able to force a misclassification of the input or even switch
the prediction to any desired class. With machine learning taking
a more important role within our society, the security of machine
learning models in general is under more scrutiny than ever. 

To define an adversarial example, we use a definition similar to \cite{ElsayedBengioEtAl:LargeMarginDeepNetworksClassification:NIPS2018_7364}.
Suppose we use a set of scoring functions $f_{j}:\mathcal{X\rightarrow}\mathbb{R}$
which assign a score to each class $j\in\mathcal{C}=\left\{ 1,\ldots,N_{c}\right\} $
given an input $\mathbf{x}$ of the data space $\mathcal{X}$. Moreover,
the predicted class label $c^{*}\left(\mathbf{x}\right)$ for $\mathbf{x}$
is determined by a winner-takes-all rule $c^{*}\left(\mathbf{x}\right)=\arg\max_{j}f_{j}\left(\mathbf{x}\right)$
and we have access to a labeled data point $\left(\mathbf{x},y\right)$
which is correctly classified as $c^{*}\left(\mathbf{x}\right)=y$.
An adversarial example $\tilde{\mathbf{x}}$ of the sample $\mathbf{x}$
is defined as the minimal required perturbation of $\mathbf{x}$ by
$\boldsymbol{\epsilon}$ to find a point at the decision boundary
or in the classification region of a different class than $y$, i.\,e.

\begin{equation}
\min_{\boldsymbol{\epsilon}}\left\Vert \boldsymbol{\epsilon}\right\Vert ,\textrm{ s.t. }f_{j}\left(\tilde{\mathbf{x}}\right)\geq f_{y}\left(\tilde{\mathbf{x}}\right)\textrm{ and }\tilde{\mathbf{x}}=\mathbf{x}+\boldsymbol{\epsilon}\in\mathcal{X}\textrm{ and \ensuremath{j\neq y.}}\label{eq:AdversarialExample}
\end{equation}
Note that the magnitude of the perturbation is measured regarding
a respective norm $\left\Vert \cdot\right\Vert $. If $f_{j}\left(\tilde{\mathbf{x}}\right)\approx f_{y}\left(\tilde{\mathbf{x}}\right)$,
an adversarial example close to the decision boundary is found. Thus,
adversarials are also related to the analysis of the decision boundaries
in a learned model. It is important to define the difference between
the ability to generalize and the robustness of a model \cite{stutz2018disentangling}.
Assume a model trained on a finite number of data points drawn from
an unknown data manifold in $\mathcal{X}$. Generalization refers
to the property to correctly classify an \emph{arbitrary} point \emph{from}
the unknown data manifold (so-called on-manifold samples). The robustness
of a model refers to the ability to correctly classify on-manifold
samples that were \emph{arbitrarily disturbed}, e.\,g. by injecting
Gaussian noise. Depending on the kind of noise these samples are on-manifold
or off-manifold adversarials (not located on the data manifold). Generalization
and robustness have to be learned explicitly because the one does
not imply the other.

Although Learning Vector Quantization (LVQ), as originally suggested
by \noun{T.~Kohonen} in \cite{kohonen88j}, is frequently claimed
as one of the most robust crisp classification approaches, its robustness
has not been actively studied yet. This claim is based on the characteristics
of LVQ methods to partition the data space into Vorono? cells (receptive
fields), according to the best matching prototype vector. For the
Generalized LVQ (GLVQ) \cite{sato96a}, considered as a differentiable
cost function based variant of LVQ, robustness is theoretically anticipated
because it maximizes the hypothesis margin in the \emph{input space}
\cite{Crammer2002a}. This changes if the squared Euclidean distance
in GLVQ is replaced by adaptive dissimilarity measures such as in
Generalized Matrix LVQ (GMLVQ) \cite{Schneider2009_MatrixLearning}
or Generalized Tangent LVQ (GTLVQ) \cite{Villmann2016h}. They first
apply a projection and measure the dissimilarity in the corresponding
\emph{projection space}, also denoted as feature space. A general
robustness assumption for these models seems to be more vague.

The \textbf{observations} of this paper are: \textbf{(1)}~GLVQ and
GTLVQ have a high robustness because of their hypothesis margin maximization
in an appropriate space. \textbf{(2)}~GMLVQ is susceptible to adversarial
attacks and hypothesis margin maximization does not guarantee a robust
model in general. \textbf{(3)}~By increasing the number of prototypes
the robustness \emph{and} the generalization ability of a LVQ model
increases. \textbf{(4)}~Adversarial examples generated for GLVQ and
GTLVQ often make semantic sense by interpolating between digits.

\section{Learning Vector Quantization\label{sec:Learning-Vector-Quantization}}

LVQ assumes a set $\mathcal{W}=\left\{ \mathbf{w}_{1},\ldots,\mathbf{w}_{N_{w}}\right\} $
of prototypes $\mathbf{w}_{k}\in\mathbb{R}^{n}$ to represent and
classify the data $\mathbf{x}\in\mathcal{X}\subseteq\mathbb{R}^{n}$
regarding a chosen dissimilarity $d\left(\mathbf{x},\mathbf{w}_{k}\right)$.
Each prototype is responsible for exactly one class $c\left(\mathbf{w}_{k}\right)\in\mathcal{C}$
and each class is represented by at least one prototype. The training
dataset is defined as a set of labeled data points $X=\left\{ \left(\mathbf{x}_{i},y_{i}\right)|\mathbf{x}_{i}\in\mathcal{X},y_{i}\in\mathcal{C}\right\} $.
The scoring function for the class $j$ yields $f_{j}\left(\mathbf{x}\right)=-\min_{k:c\left(\mathbf{w}_{k}\right)=j}d\left(\mathbf{x},\mathbf{w}_{k}\right)$.
Hence, the predicted class $c^{*}\left(\mathbf{x}\right)$ is the
class label $c\left(\mathbf{w}_{k}\right)$ of the closest prototype
$\mathbf{w}_{k}$ to $\mathbf{x}$.

\subsubsection*{Generalized LVQ:}

GLVQ is a cost function based variant of LVQ such that stochastic
gradient descent learning (SGDL) can be performed as optimization
strategy \cite{sato96a}. Given a training sample $\left(\mathbf{x}_{i},y_{i}\right)\in X$,
the two \emph{closest} prototypes $\mathbf{w}^{+}\in\mathcal{W}$
and $\mathbf{w}^{-}\in\mathcal{W}$ with correct label $c\left(\mathbf{w}^{+}\right)=y_{i}$
and incorrect label $c\left(\mathbf{w}^{-}\right)\neq y_{i}$ are
determined. The dissimilarity function is defined as the squared Euclidean
distance $d_{E}^{2}\left(\mathbf{x},\mathbf{w}_{k}\right)=\left(\mathbf{x}-\mathbf{w}_{k}\right)^{T}\left(\mathbf{x}-\mathbf{w}_{k}\right)$.
The cost function of GLVQ is
\begin{equation}
E_{GLVQ}\left(X,\mathcal{W}\right)=\sum_{\left(\mathbf{x}_{i},y_{i}\right)\in X}l\left(\mathbf{x}_{i},y_{i},\mathcal{W}\right)\label{eq:cost GLVQ}
\end{equation}
with the local loss $l\left(\mathbf{x}_{i},y_{i},\mathcal{W}\right)=\varphi\left(\mu\left(\mathbf{x}_{i},y_{i},\mathcal{W}\right)\right)$
where $\varphi$ is a monotonically increasing differentiable activation
function. The classifier function $\mu$ is defined as
\begin{equation}
\mu\left(\mathbf{x}_{i},y_{i},\mathcal{W}\right)=\frac{d^{+}\left(\mathbf{x}_{i}\right)-d^{-}\left(\mathbf{x}_{i}\right)}{d^{+}\left(\mathbf{x}_{i}\right)+d^{-}\left(\mathbf{x}_{i}\right)}\in\left[-1,1\right]\label{eq:classifier function}
\end{equation}
where $d^{\pm}\left(\mathbf{x}_{i}\right)=d_{E}^{2}\left(\mathbf{x}_{i},\mathbf{w}^{\pm}\right)$.
Thus, $\mu\left(\mathbf{x}_{i},y_{i},\mathcal{W}\right)$ is negative
for a correctly classified training sample $\left(\mathbf{x}_{i},y_{i}\right)$
and positive otherwise. Since $l\left(\mathbf{x}_{i},y_{i},\mathcal{W}\right)$
is differentiable, the prototypes $\mathcal{W}$ can be learned by
a SGDL approach.

\subsubsection*{Generalized Matrix LVQ:}

By substituting the dissimilarity measure $d_{E}^{2}$ in GLVQ with
an adaptive dissimilarity measure
\begin{equation}
d_{\boldsymbol{\Omega}}^{2}\left(\mathbf{x},\mathbf{w}_{k}\right)=d_{E}^{2}\left(\boldsymbol{\Omega}\mathbf{x},\boldsymbol{\Omega}\mathbf{w}_{k}\right),\label{eq:GMLVQ}
\end{equation}
GMLVQ is obtained \cite{Schneider2009_MatrixLearning}. The relevance
matrix $\boldsymbol{\Omega}\in\mathbb{R}^{r\times n}$ is learned
during training in parallel to the prototypes. The parameter $r$
controls the projection dimension of $\boldsymbol{\Omega}$ and must
be defined in advance.

\subsubsection*{Generalized Tangent LVQ:}

In contrast to the previous methods, GTLVQ \cite{Villmann2016h} defines
the prototypes as affine subspaces in $\mathbb{R}^{n}$ instead of
points. More precisely, the set of prototypes is defined as $\mathcal{W}_{T}=\left\{ \left(\mathbf{w}_{1},\mathbf{W}_{1}\right),\ldots,\left(\mathbf{w}_{N_{w}},\mathbf{W}_{N_{w}}\right)\right\} $
where $\mathbf{W}_{k}\in\mathbb{R}^{n\times r}$ is the $r$-dimensional
basis and $\mathbf{w}_{k}$ is the translation vector of the affine
subspace. Together with the parameter vector $\boldsymbol{\theta}\in\mathbb{R}^{r}$,
they form the prototype as affine subspace $\mathbf{w}_{k}+\mathbf{W}_{k}\boldsymbol{\theta}$.
The tangent distance is defined as

\begin{equation}
d_{T}^{2}\left(\mathbf{x},\left(\mathbf{w}_{k},\mathbf{W}_{k}\right)\right)=\min_{\boldsymbol{\theta}\in\mathbb{R}^{r}}d_{E}^{2}\left(\mathbf{x},\mathbf{w}_{k}+\mathbf{W}_{k}\boldsymbol{\theta}\right)\label{eq:GTLVQ}
\end{equation}
where $r$ is a hyperparameter. Substituting $d_{E}^{2}$ in GLVQ
with $d_{T}^{2}$ and redefining the set of prototypes to $\mathcal{W}_{T}$
yields GTLVQ. The affine subspaces defined by $\left(\mathbf{w}_{k},\mathbf{W}_{k}\right)$
are learned by SGDL.

\section{Experimental Setup}

In this section adversarial attacks as well as robustness metrics
are introduced and the setup of the evaluation is explained. The setup
used here follows the one presented in \cite{SchottEtAl:AdversariallyRobustNNforMNIST:2018}
with a few minor modifications to the study of LVQ methods. All experiments
and models were implemented using the \noun{Keras} framework in \noun{Python}
on top of \noun{Tensorflow}.\footnote{\noun{Tensorflow}: \url{www.tensorflow.org}; \noun{Keras:} \url{www.keras.io}}
All evaluated LVQ models are made available as pretrained \noun{Tensorflow
}graphs and as part of the \noun{Foolbox zoo}\footnote{\url{https://foolbox.readthedocs.io/en/latest/modules/zoo.html}}
at \url{https://github.com/LarsHoldijk/robust_LVQ_models}\noun{.} 

The \noun{Foolbox} \cite{RauberEtAl:FoolboxPythonToolboxBenchmarkAdversarialRobustness:ICML2017}
implementations with default settings were used for the attacks. The
evaluation was performed using the MNIST dataset as it is one of the
most used datasets for robust model evaluation in the literature.
Despite being considered by many as a solved `toy' dataset with state-of-the-art
(SOTA) deep learning models reaching close to perfect classification
accuracy, the defense of adversarial attacks on MNIST is still far
from being trivial \cite{SchottEtAl:AdversariallyRobustNNforMNIST:2018}.
The dataset consists of handwritten digits in the data space $\mathcal{X=}\left[0,1\right]^{n}$
with $n=28\cdot28$. We trained our models on the 60K training images
and evaluated all metrics and scores on the \emph{complete} 10K test
images.

\subsection{Adversarial Attacks\label{subsec:Adversarial-Attacks}}

Adversarial attacks can be grouped into two different approaches,
white-box and black-box, distinguished by the amount of knowledge
about the model available to the attacker. White-box or gradient-based
attacks are based on exploiting the interior gradients of the NNs,
while black-box attacks rely only on the output of the model, either
the logits, the probabilities or just the predicted discrete class
labels. Each attack is designed to optimize the adversarial image
regarding a given norm. Usually, the attacks are defined to optimize
over $L^{p}$ norms (or $p$-norms) with $p\in\left\{ 0,2,\infty\right\} $
and, therefore, are called $L^{p}$-attacks.

In the evaluation, nine attacks including white-box and black-box
attacks were compared. The white-box attacks are: Fast Gradient Sign
Method (FGSM) \cite{GoodfellowEtAl:ExplainingAndHarnessingAdversarialExamples:arXiv2015},
Fast Gradient Method (FGM), Basic Iterative Method (BIM) \cite{KurakinEtAl:AdversarialExamplesPhysicalWorld:arXiv2017},
Momentum Iterative Method (MIM) \cite{dong2018boosting} and Deepfool
\cite{Moosavi-Dezfooli2015}. The black-box attacks are: Gaussian
blur, Salt-and-Pepper (S\&P), Pointwise \cite{SchottEtAl:AdversariallyRobustNNforMNIST:2018}
and Boundary \cite{BrendelEtAl:DecisionBasedAdversarialAttacksforBlackBoxMachineLearningModels:Proc.ICLR2018}.
See Tab.~\ref{tab:robustness_comparison} for the $L^{p}$ definition
of each attack. Note that some of the attacks are defined for more
than one norm.

\subsection{Robustness Metrics}

The robustness of a model is measured by four different metrics, all
based on the \emph{adversarial distances} $\delta_{A}\left(\mathbf{x},y\right)$.
Given a labeled test sample $\left(\mathbf{x},y\right)$ from a test
set $T$ and an adversarial $L^{p}$-attack $A$, $\delta_{A}\left(\mathbf{x},y\right)$
is defined as: \textbf{(1)}~zero if the data sample is misclassified
$c^{*}\left(\mathbf{x}\right)\neq y$; \textbf{(2)}~$\left\Vert \boldsymbol{\epsilon}\right\Vert _{p}=\left\Vert \tilde{\mathbf{x}}-\mathbf{x}\right\Vert _{p}$
if $A$ found an adversary $\tilde{\mathbf{x}}$ and $c^{*}\left(\mathbf{x}\right)=y$;
\textbf{(3)}~$\infty$ if no adversary was found by $A$ and $c^{*}\left(\mathbf{x}\right)=y$. 

For each attack $A$ the \emph{median-$\delta_{A}$} score is defined
as $\textrm{median}\left\{ \delta_{A}\left(\mathbf{x},y\right)|\left(\mathbf{x},y\right)\in T\right\} ,$
describing an averaged $\delta_{A}$ over $T$ robust to outliers.\footnote{Hence, \emph{median-$\delta_{A}$} can be $\infty$ if for over $50\%$
of the samples no adversary was found.} The \emph{median-$\delta_{p}^{*}$} score is computed for all $L^{p}$-attacks
as the $\textrm{median}\left\{ \delta_{p}^{*}\left(\mathbf{x},y\right)|\left(\mathbf{x},y\right)\in T\right\} $
where $\delta_{p}^{*}\left(\mathbf{x},y\right)$ is defined as $\min\left\{ \delta_{A}\left(\mathbf{x},y\right)|A\text{\textrm{ is a }}L^{p}\textrm{-attack}\right\} $.
This score is a worst-case evaluation of the median-\emph{$\delta_{A}$,}
assuming that each sample is disturbed by the respective worst-case
attack $A_{p}^{*}$ (the attack with the smallest distance). Additionally,
the threshold accuracies \emph{acc-}$A$ and \emph{acc-}$A_{p}^{*}$
of a model over $T$ are defined as the percentage of adversarial
examples found with $\delta_{A}\left(\mathbf{x},y\right)\leq t_{p}$,
using either the given $L^{p}$-attack $A$ for all samples or the
respective worst-case attack $A_{p}^{*}$ respectively. This metric
represents the remaining accuracy of the model when only adversaries
under a given threshold are considered valid. We used the following
thresholds for our evaluation: $t_{0}=12$, $t_{2}=1.5$ and $t_{\infty}=0.3$.

\subsection{Training Setup and Models}

\begin{table}[t]
\caption{The results of the robustness evaluation. Attacks are clustered by
their $L^{p}$ class, the boxes denote the type of the attack (white-
or black-box). Accuracies are given in percentages and the \#prototypes
is recorded per class. All scores are evaluated on the test set. For
each model we report the clean accuracy (clean acc.), the median-$\delta_{A}$
(left value) and acc-$A$ score (right value) for each attack and
the worst-case (worst-c.) analysis over all $L^{p}$-attacks by presenting
the median-$\delta_{p}^{*}$ (left value) and acc-$A_{p}^{*}$ score
(right value). Higher scores mean higher robustness of the model.
The median-$\delta_{A}$ of the most robust model in each attack is
highlighted in bold. Overall, the model with the best (highest) worst-case
median-$\delta_{p}^{*}$ is underlined and highlighted.\label{tab:robustness_comparison}}

\centering{}{\scriptsize{}}%
\begin{longtable}{ll|>{\centering}p{0.5cm}>{\centering}p{0.5cm}||>{\centering}p{0.5cm}>{\centering}p{0.5cm}||>{\centering}p{0.5cm}>{\centering}p{0.5cm}|>{\centering}p{0.5cm}>{\centering}p{0.5cm}||>{\centering}p{0.5cm}>{\centering}p{0.5cm}|>{\centering}p{0.5cm}>{\centering}p{0.5cm}||>{\centering}p{0.5cm}>{\centering}p{0.5cm}|>{\centering}p{0.5cm}>{\centering}p{0.5cm}}
 &  & \multicolumn{2}{c||}{{\footnotesize{}CNN}} & \multicolumn{2}{c||}{{\footnotesize{}Madry}} & \multicolumn{4}{c||}{{\footnotesize{}GLVQ}} & \multicolumn{4}{c||}{{\footnotesize{}GMLVQ}} & \multicolumn{4}{c}{{\footnotesize{}GTLVQ}}\tabularnewline
\hline 
\multicolumn{2}{l|}{{\footnotesize{}\#prototypes}} &  &  &  &  & \multicolumn{2}{c|}{{\footnotesize{}1}} & \multicolumn{2}{c||}{{\footnotesize{}128}} & \multicolumn{2}{c|}{{\footnotesize{}1}} & \multicolumn{2}{c||}{{\footnotesize{}49}} & \multicolumn{2}{c|}{{\footnotesize{}1}} & \multicolumn{2}{c}{{\footnotesize{}10}}\tabularnewline
\hline 
\multicolumn{2}{l|}{{\footnotesize{}Clean acc.}} &  & {\footnotesize{}99} &  & {\footnotesize{}99} &  & {\footnotesize{}83} &  & {\footnotesize{}95} &  & {\footnotesize{}88} &  & {\footnotesize{}93} &  & {\footnotesize{}95} &  & {\footnotesize{}97}\tabularnewline
\hline 
\multirow{7}{*}{\begin{turn}{90}
{\footnotesize{}$L^{2}$}
\end{turn}} & {\footnotesize{}FGM} \hspace*{\fill}{\tiny{}$\square$} & {\footnotesize{}2.1} & \textcolor{black}{\footnotesize{}73} & \textbf{\footnotesize{}$\boldsymbol{\infty}$} & {\footnotesize{}96} & {\footnotesize{}$\infty$} & {\footnotesize{}63} & {\footnotesize{}$\infty$} & {\footnotesize{}76} & {\footnotesize{}0.6} & {\footnotesize{}7} & {\footnotesize{}0.8} & {\footnotesize{}15} & {\footnotesize{}$\infty$} & {\footnotesize{}71} & {\footnotesize{}$\infty$} & {\footnotesize{}81}\tabularnewline
 & {\footnotesize{}Deepfool} \hspace*{\fill}{\tiny{}$\square$} & {\footnotesize{}1.9} & \textcolor{black}{\footnotesize{}70} & \textbf{\footnotesize{}5.5} & {\footnotesize{}94} & {\footnotesize{}1.6} & {\footnotesize{}53} & {\footnotesize{}2.3} & {\footnotesize{}73} & {\footnotesize{}0.5} & {\footnotesize{}26} & {\footnotesize{}0.7} & {\footnotesize{}27} & {\footnotesize{}2.3} & {\footnotesize{}73} & {\footnotesize{}2.5} & {\footnotesize{}81}\tabularnewline
 & {\footnotesize{}BIM} \hspace*{\fill}{\tiny{}$\square$} & {\footnotesize{}1.5} & \textcolor{black}{\footnotesize{}50} & \textbf{\footnotesize{}4.9} & {\footnotesize{}94} & {\footnotesize{}1.5} & {\footnotesize{}50} & {\footnotesize{}2.1} & {\footnotesize{}68} & {\footnotesize{}0.6} & {\footnotesize{}6} & {\footnotesize{}0.7} & {\footnotesize{}8} & {\footnotesize{}2.1} & {\footnotesize{}68} & {\footnotesize{}2.3} & {\footnotesize{}77}\tabularnewline
 & {\footnotesize{}Gaussian} \hspace*{\fill}{\tiny{}$\blacksquare$} & {\footnotesize{}6.4} & \textcolor{black}{\footnotesize{}99} & {\footnotesize{}6.6} & {\footnotesize{}98} & {\footnotesize{}6.8} & {\footnotesize{}83} & {\footnotesize{}6.7} & {\footnotesize{}68} & {\footnotesize{}6.3} & {\footnotesize{}88} & {\footnotesize{}6.2} & {\footnotesize{}92} & \textbf{\footnotesize{}7.1} & {\footnotesize{}94} & {\footnotesize{}6.9} & {\footnotesize{}97}\tabularnewline
 & {\footnotesize{}Pointwise} \hspace*{\fill}{\tiny{}$\blacksquare$} & {\footnotesize{}4.2} & \textcolor{black}{\footnotesize{}96} & {\footnotesize{}2.1} & {\footnotesize{}80} & {\footnotesize{}4.5} & {\footnotesize{}79} & {\footnotesize{}5.4} & {\footnotesize{}92} & {\footnotesize{}1.6} & {\footnotesize{}54} & {\footnotesize{}2.4} & {\footnotesize{}78} & {\footnotesize{}5.5} & {\footnotesize{}92} & \textbf{\footnotesize{}5.6} & {\footnotesize{}95}\tabularnewline
 & {\footnotesize{}Boundary} \hspace*{\fill}{\tiny{}$\blacksquare$} & {\footnotesize{}1.9} & \textcolor{black}{\footnotesize{}76} & {\footnotesize{}1.5} & {\footnotesize{}52} & {\footnotesize{}2.1} & {\footnotesize{}61} & \textbf{\footnotesize{}3.2} & {\footnotesize{}76} & {\footnotesize{}0.6} & {\footnotesize{}7} & {\footnotesize{}0.8} & {\footnotesize{}7} & {\footnotesize{}2.8} & {\footnotesize{}78} & {\footnotesize{}3.1} & {\footnotesize{}86}\tabularnewline
\cline{2-18} 
 & \textbf{\footnotesize{}worst-c.} & {\footnotesize{}1.5} & \textcolor{black}{\footnotesize{}50} & {\footnotesize{}1.5} & {\footnotesize{}52} & {\footnotesize{}1.5} & {\footnotesize{}49} & {\footnotesize{}2.1} & {\footnotesize{}68} & {\footnotesize{}0.5} & {\footnotesize{}3} & {\footnotesize{}0.6} & {\footnotesize{}3} & {\footnotesize{}2.1} & {\footnotesize{}68} & \textbf{\footnotesize{}\uline{2.2}} & {\footnotesize{}77}\tabularnewline
\hline 
\hline 
\multirow{5}{*}{\begin{turn}{90}
{\footnotesize{}$L^{\infty}$}
\end{turn}} & {\footnotesize{}FGSM} \hspace*{\fill}{\tiny{}$\square$} & {\footnotesize{}.17} & \textcolor{black}{\footnotesize{}7} & \textbf{\footnotesize{}.52} & {\footnotesize{}96} & {\footnotesize{}.17} & {\footnotesize{}11} & {\footnotesize{}.29} & {\footnotesize{}43} & {\footnotesize{}.04} & {\footnotesize{}0} & {\footnotesize{}.05} & {\footnotesize{}0} & {\footnotesize{}.22} & {\footnotesize{}18} & {\footnotesize{}.25} & {\footnotesize{}26}\tabularnewline
 & {\footnotesize{}Deepfool} \hspace*{\fill}{\tiny{}$\square$} & {\footnotesize{}.16} & \textcolor{black}{\footnotesize{}1} & \textbf{\footnotesize{}.49} & {\footnotesize{}95} & {\footnotesize{}.13} & {\footnotesize{}7} & {\footnotesize{}.22} & {\footnotesize{}21} & {\footnotesize{}.04} & {\footnotesize{}27} & {\footnotesize{}.05} & {\footnotesize{}19} & {\footnotesize{}.19} & {\footnotesize{}9} & {\footnotesize{}.22} & {\footnotesize{}19}\tabularnewline
 & {\footnotesize{}BIM} \hspace*{\fill}{\tiny{}$\square$} & {\footnotesize{}.12} & \textcolor{black}{\footnotesize{}0} & \textbf{\footnotesize{}.41} & {\footnotesize{}94} & {\footnotesize{}.12} & {\footnotesize{}3} & {\footnotesize{}.20} & {\footnotesize{}9} & {\footnotesize{}.04} & {\footnotesize{}0} & {\footnotesize{}.05} & {\footnotesize{}0} & {\footnotesize{}.17} & {\footnotesize{}3} & {\footnotesize{}.20} & {\footnotesize{}5}\tabularnewline
 & \textcolor{black}{\footnotesize{}MIM} \hspace*{\fill}{\tiny{}$\square$} & \textcolor{black}{\footnotesize{}.13} & \textcolor{black}{\footnotesize{}0} & \textbf{\textcolor{black}{\footnotesize{}.38}} & \textcolor{black}{\footnotesize{}93} & \textcolor{black}{\footnotesize{}.12} & \textcolor{black}{\footnotesize{}3} & {\footnotesize{}.19} & {\footnotesize{}9} & \textcolor{black}{\footnotesize{}.04} & \textcolor{black}{\footnotesize{}0} & {\footnotesize{}.05} & {\footnotesize{}0} & \textcolor{black}{\footnotesize{}.17} & \textcolor{black}{\footnotesize{}3} & {\footnotesize{}.20} & {\footnotesize{}5}\tabularnewline
\cline{2-18} 
 & \textbf{\footnotesize{}worst-c.} & {\footnotesize{}.12} & \textcolor{black}{\footnotesize{}0} & \textbf{\footnotesize{}\uline{.38}} & {\footnotesize{}93} & {\footnotesize{}.11} & {\footnotesize{}2} & {\footnotesize{}.19} & {\footnotesize{}5} & {\footnotesize{}.03} & {\footnotesize{}0} & {\footnotesize{}.04} & {\footnotesize{}0} & {\footnotesize{}.17} & {\footnotesize{}3} & {\footnotesize{}.19} & {\footnotesize{}4}\tabularnewline
\hline 
\hline 
\multirow{3}{*}{\begin{turn}{90}
{\footnotesize{}$L^{0}$}
\end{turn}} & {\footnotesize{}Pointwise} \hspace*{\fill}{\tiny{}$\blacksquare$} & {\footnotesize{}19} & \textcolor{black}{\footnotesize{}73} & {\footnotesize{}4} & {\footnotesize{}1} & {\footnotesize{}22} & {\footnotesize{}64} & {\footnotesize{}32} & {\footnotesize{}79} & {\footnotesize{}3} & {\footnotesize{}6} & {\footnotesize{}6} & {\footnotesize{}18} & {\footnotesize{}34} & {\footnotesize{}80} & \textbf{\footnotesize{}35} & {\footnotesize{}85}\tabularnewline
 & {\footnotesize{}S\&P} \hspace*{\fill}{\tiny{}$\blacksquare$} & {\footnotesize{}65} & \textcolor{black}{\footnotesize{}94} & {\footnotesize{}17} & {\footnotesize{}63} & {\footnotesize{}126} & {\footnotesize{}77} & \textbf{\footnotesize{}188} & {\footnotesize{}92} & {\footnotesize{}8} & {\footnotesize{}37} & {\footnotesize{}17} & {\footnotesize{}61} & {\footnotesize{}155} & {\footnotesize{}91} & {\footnotesize{}179} & {\footnotesize{}95}\tabularnewline
\cline{2-18} 
 & \textbf{\footnotesize{}worst-c.} & {\footnotesize{}19} & \textcolor{black}{\footnotesize{}73} & {\footnotesize{}4} & {\footnotesize{}1} & {\footnotesize{}22} & {\footnotesize{}64} & {\footnotesize{}32} & {\footnotesize{}79} & {\footnotesize{}3} & {\footnotesize{}6} & {\footnotesize{}6} & {\footnotesize{}18} & {\footnotesize{}34} & {\footnotesize{}80} & \textbf{\footnotesize{}\uline{35}} & {\footnotesize{}85}\tabularnewline
\end{longtable}{\scriptsize\par}
\end{table}
All models, except the Madry model, were trained with the Adam optimizer
\cite{KingmaBa:AdamMethodStochasticGradient:ICLR2015} for 150 epochs
using basic data augmentation in the form of random shifts by $\pm2$\,pixels
and random rotations by $\pm15^{{^\circ}}$.

\subsubsection*{NN Models:\label{subsec:Baseline-models}}

Two NNs are used as baseline models for the evaluation. The first
model is a convolutional NN, denoted as CNN, with two convolutional
layers and two fully connected layers. The convolutional layers have
32 and 64 filters with a stride of one and a kernel size of 3$\times$3.
Both are followed by max-pooling layers with a window size and stride
each of 2$\times$2. None of the layers use padding. The first fully
connected layer has 128 neurons and a dropout rate of 0.5. All layers
use the ReLU activation function except for the final fully connected
output layer which uses a softmax function. The network was trained
using the categorical cross entropy loss and an initial learning rate
of $10^{-4}$ with a decay of 0.9 at plateaus. 

The second baseline model is the current SOTA model for MNIST in terms
of robustness proposed in \cite{Madry2017} and denoted as Madry.
This model relies on a special kind of adversarial training by considering
it as a min-max optimization game: before the loss function is minimized
over a given training batch, the original images are partially substituted
by perturbed images with $\left\Vert \boldsymbol{\epsilon}\right\Vert _{\infty}\leq0.3$
such that the loss function is \emph{maximized} over the given batch.
The Madry model was downloaded from \url{https://github.com/MadryLab/mnist_challenge}.

\subsubsection*{LVQ Models:}

All three LVQ models were trained using an initial learning rate of
0.01 with a decay of 0.5 at plateaus and with $\varphi$ defined as
the identity function. The prototypes (translation vectors) of all
methods were class-wise initialized by k-means over the training dataset.
For GMLVQ, we defined $\boldsymbol{\Omega}$ with $n=r$ and initialized
$\boldsymbol{\Omega}$ as a scaled identity matrix with Frobenius
norm one. After each update step, $\boldsymbol{\Omega}$ was normalized
to again have Frobenius norm one. The basis matrices $\mathbf{W}_{k}$
of GTLVQ were defined by $r=12$ and initialized by a singular value
decomposition with respect to each initialized prototype $\mathbf{w}_{k}$
over the set of class corresponding training points \cite{Villmann2016h}.
The prototypes were not constrained to $\mathcal{X}$ (`box constrained')
during the training, resulting in possibly non-interpretable prototypes
as they can be points in $\mathbb{R}^{n}$.\footnote{A restriction to $\mathcal{X}$ leads to an accuracy decrease of less
than $1\%$.}

Two versions of each LVQ model were trained: one with one prototype
per class and one with multiple prototypes per class. For the latter
the numbers of prototypes were chosen such that all LVQ models have
roughly 1M parameters. The chosen number of prototypes per class are
given in Tab.~\ref{tab:robustness_comparison} by \#prototypes.

\section{Results}

\begin{figure}
\begin{centering}
\begin{tabular}{lcccccccccc}
CNN & \includegraphics{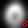} & \includegraphics{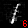} & \includegraphics{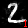} & \includegraphics{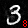} & \includegraphics{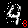} & \includegraphics{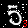} & \includegraphics{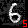} & \includegraphics{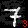} & \includegraphics{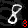} & \includegraphics{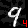}\tabularnewline
Madry & \includegraphics{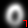} & \includegraphics{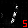} & \includegraphics{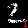} & \includegraphics{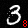} & \includegraphics{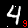} & \includegraphics{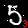} & \includegraphics{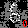} & \includegraphics{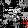} & \includegraphics{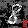} & \includegraphics{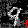}\tabularnewline
GLVQ & \includegraphics{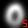} & \includegraphics{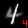} & \includegraphics{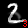} & \includegraphics{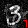} & \includegraphics{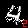} & \includegraphics{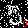} & \includegraphics{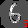} & \includegraphics{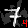} & \includegraphics{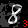} & \includegraphics{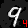}\tabularnewline
GMLVQ & \includegraphics{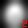} & \includegraphics{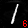} & \includegraphics{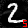} & \includegraphics{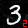} & \includegraphics{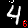} & \includegraphics{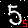} & \includegraphics{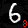} & \includegraphics{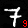} & \includegraphics{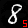} & \includegraphics{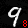}\tabularnewline
GTLVQ & \includegraphics{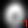} & \includegraphics{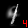} & \includegraphics{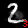} & \includegraphics{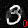} & \includegraphics{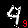} & \includegraphics{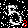} & \includegraphics{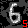} & \includegraphics{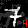} & \includegraphics{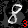} & \includegraphics{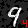}\tabularnewline
\end{tabular}
\par\end{centering}
\caption{For each model, adversarial examples generated by the attacks (from
left to right): Gaussian, Deepfool ($L_{2}$), BIM ($L_{2}$), Boundary,
Pointwise ($L_{0}$), S\&P, FGSM, Deepfool ($L_{\infty}$), BIM ($L_{\infty}$)
and MIM. For the LVQ models the version with more prototypes per class
was used. The ten digits were randomly selected under the condition
that every digit was classified correctly by all models. The original
images are 0, 1, ..., 9 from left to right. The red digits in the
lower right corners indicate the models prediction after the adversarial
attack.\label{fig: images}}
\end{figure}
The results of the model robustness evaluation are presented in Tab.~\ref{tab:robustness_comparison}.
Fig.~\ref{fig: images} displays adversarial examples generated for
each model. Below, the four most notable observations that can be
made from the results are discussed.

\subsubsection*{Hypothesis margin maximization in the input space produces robust
models (GLVQ and GTLVQ are highly robust): \label{subsec:Hypothesis-margin-maximization in the input space}}

Tab.~\ref{tab:robustness_comparison} shows outstanding robustness
against adversarial attacks for GLVQ and GTLVQ. GLVQ with multiple
prototypes and GTLVQ with both one or more prototypes per class, outperform
the NN models by a large difference for the $L^{0}$- and $L^{2}$-attacks
while having a considerably lower clean accuracy. This is not only
the case for individual black-box attacks but also for the worst-case
scenarios. For the $L^{0}$-attacks this difference is especially
apparent. A possible explanation is that the robustness of GLVQ and
GTLVQ is achieved due to the input space hypothesis margin maximization
\cite{Crammer2002a}.\footnote{Note that the results of \cite{Crammer2002a} hold for GTLVQ as it
can be seen as a version of GLVQ with infinitely many prototypes learning
the affine subspaces.} In \cite{Crammer2002a} it was stated that the hypothesis margin
is a lower bound for the sample margin which is, \emph{if defined
in the input space}, used in the definition of adversarial examples
(\ref{eq:AdversarialExample}). \emph{Hence, if we maximize the hypothesis
margin in the input space we guarantee a high sample margin and therefore,
a robust model.} A first attempt to transfer this principle was made
in \cite{ElsayedBengioEtAl:LargeMarginDeepNetworksClassification:NIPS2018_7364}
to create a robust NN by a first order approximation of the sample
margin in the input space.

However, the Madry model still outperforms GLVQ and GTLVQ in the $L^{\infty}$-attacks
as expected. This result is easily explained using the manifold based
definition of adversarial examples and the adversarial training procedure
of the Madry model, which optimizes the robustness against $\left\Vert \boldsymbol{\epsilon}\right\Vert _{\infty}\leq0.3$.
Considering the manifold definition, one could say that Madry augmented
the original \noun{MNIST} manifold to include small $L^{\infty}$
perturbations. Doing so, Madry creates a new \emph{training}-manifold
in addition to the original \noun{MNIST} manifold. In other words,
the $L^{\infty}$ robustness of the adversarial trained Madry model
can be seen as its generalization on the new training-manifold (this
becomes clear if one considers the high acc-$A$ scores for $L^{\infty}$).
For this reason, the Madry model is only robust against off-manifold
examples that are on the generated training-manifold. As soon as off-training-manifold
examples are considered the accuracy will drop fast. This was also
shown in \cite{SchottEtAl:AdversariallyRobustNNforMNIST:2018}, where
the accuracy of the Madry model is significantly lower when considering
a threshold $t_{\infty}>0.3$.\footnote{For future work a more extensive evaluation should be considered:
including not only the norm for which a single attack was optimized
but rather a combination of all three norms. This gives a better insight
on the characteristics of the attack and the defending model. The
$L^{0}$ norm can be interpreted as the number of pixels that have
to change, the $L^{\infty}$ norm as the maximum deviation of a pixel
and the $L^{2}$ norm as a kind of average pixel change. As attacks
are optimized for a certain norm, only considering this norm might
give a skewed impression of their attacking capability. Continuing,
calculating a threshold accuracy including only adversaries that are
below all three thresholds may give an interesting and more meaningful
metric.}

Furthermore, the Madry model has outstanding robustness scores for
gradient-based attacks in general. We accredit this effect to potential
obfuscation of gradients as a side-effect of the adversarial training
procedure. While \cite{athalye2018obfuscated} was not able to find
concrete evidence of gradient obfuscation due to adversarial training
in the Madry model, it did list black-box-attacks outperforming white-box
attacks as a signal for its occurrence.

\subsubsection*{Hypothesis margin maximization in a space different to the input
space does not necessarily produce robust models (GMLVQ is susceptible
for adversarial attacks):}

In contrast to GLVQ and GTLVQ, GMLVQ has the lowest robustness score
across all attacks and all methods. Taking the strong relation of
GTLVQ and GMLVQ into account \cite{Villmann2016h}, it is a remarkable
result.\footnote{GTLVQ can be seen as localized version of GMLVQ with the constraint
that the $\boldsymbol{\Omega}$ matrices must be orthogonal projectors. } One potential reason is, that GMLVQ maximizes the hypothesis margin
in a projection space which differ in general from the input space.
The margin maximization in the projection space is used to construct
a model with good generalization abilities, which is why GMLVQ usually
outperforms GLVQ in terms of accuracy (see the clean accuracy for
GLVQ and GMLVQ with one prototype per class). However, a large margin
in the projection space does not guarantee a big margin in the input
space. Thus, GMLVQ does not implicitly optimize the separation margin,
as used in the definition of an adversarial example (\ref{eq:AdversarialExample}),
in the input space. Hence, GMLVQ is a good example to show that a
model, which generalizes well, is not necessarily robust.

Another effect which describes the observed lack of robustness by
GMLVQ is its tendency to oversimplify (to collapse data dimensions)
without regularization. Oversimplification may induce heavy distortions
in the mapping between input and projection space, potentially creating
dimensions in which a small perturbation in the input space can be
mapped to a large perturbation in the projection space. These dimensions
are later used to efficiently place the adversarial attack. This effect
is closely related to theory known from metric learning, here oversimplification
was used by \cite{globerson2006metric} to optimize a classifier over
$d_{\boldsymbol{\Omega}}^{2}$, which \emph{maximally collapses (concentrates)
the classes to single points} (related to the prototypes in GMLVQ).
It is empirically shown that this effect helps to achieve a good generalization. 

To improve the robustness of GMLVQ penalizing the collapsing of dimensions
may be a successful approach. A method to achieve this is to force
the eigenvalue spectrum of the mapping to follow a uniform distribution,
as proposed in \cite{Villmann2010l}. This regularization technique
would also strengthen the transferability between the margin in the
projection and input space. Unfortunately, it requires the possibly
numerical instable computation of the derivative of a determinant
of a product of $\boldsymbol{\Omega}$ which makes it impossible to
train an appropriate model for MNIST using this regularization so
far. The fact that GTLVQ is a constrained version of GMLVQ gives additional
reason to believe that regularizations\,/\,constraints are able
to force a model to be more robust.

\subsubsection*{Increasing the number of prototypes improves the ability to generalize
and the robustness:}

For all three LVQ models the robustness improves if the number of
prototypes per class increases. Additionally, increasing the number
of prototypes leads to a better ability to generalize. This observation
provides empirical evidence supporting the results of \cite{stutz2018disentangling}.
In \cite{stutz2018disentangling} it was stated that generalization
and robustness are not necessarily contradicting goals, which is a
topic recently under discussion.

With multiple prototypes per class, the robustness of the GLVQ model
improves by a significantly larger margin than GTLVQ. This can be
explained by the high accuracy of GTLVQ with one prototype. The high
accuracy with one prototype per class indicates that the data manifold
of \noun{MNIST} is almost flat and can therefore be described with
one tangent such that introducing more prototypes does not improve
the model's generalization ability. If we add more prototypes in GLVQ,
the prototypes will start to approximate the data manifold and with
that implicitly the tangent prototypes used in GTLVQ. With more prototypes
per class, the scores of GLVQ will therefore most likely converge
towards those of GTLVQ.

\subsubsection*{GLVQ and GTLVQ require semantically correct adversarial examples:}

Fig.~\ref{fig: images} shows a large semantic difference between
the adversarial examples generated for GLVQ\,/\,GTLVQ and the other
models. A large portion of the adversarial examples generated for
the GLVQ and GTLVQ models look like interpolations between the original
digit and another digit.\footnote{A similar effect was observed in \cite{SchottEtAl:AdversariallyRobustNNforMNIST:2018}
for k-NN models.} This effect is especially visible for the Deepfool, BIM and Boundary
attacks. In addition to this, the Pointwise attack is required to
generate features from other digits to fool the models, e.\,g. the
horizontal bar of a two in the case of GLVQ and the closed ring of
a nine for GTLVQ (see digit four). In other words, for GLVQ and GTLVQ
some of the attacks generate adversaries that closer resemble on-manifold
samples than off-manifold. For the other models, the adversaries are
more like off-manifold samples (or in the case of Madry, off-training-manifold).

\section{Conclusion}

In this paper we extensively evaluated the robustness of LVQ models
against adversarial attacks. Most notably, we have shown that there
is a large difference in the robustness of the different LVQ models,
even if they all perform a hypothesis margin maximization. GLVQ and
GTLVQ show high robustness against adversarial attacks, while GMLVQ
scores the lowest across all attacks and all models. The discussion
related to this observation has lead to four important \textbf{conclusions}:\emph{
}\textbf{(1)}~For (hypothesis) margin maximization to lead to robust
models the space in which the margin is maximized matters, this must
be the same space as where the attack is placed. \textbf{(2)}~Collapsed
dimensions are beneficial for the generalization ability of a model.
However, they can be harmful for the model's robustness. \textbf{(3)}~It
is possible to derive a robust model by applying a fitting regularization\,/\,constraint.
This can be seen in the relation between GTLVQ and GMLVQ and is also
studied for NNs \cite{croce2018provable}. \textbf{(4)}~Our experimental
results with an increased number of prototypes support the claim of
\cite{stutz2018disentangling}, that the ability to generalize and
the robustness are principally not contradicting goals.

In summary, the overall robustness of LVQ models is impressive. Using
only one prototype per class and no purposefully designed adversarial
training, GTLVQ is on par with SOTA robustness on MNIST. With further
research, the robustness of LVQ models against adversarial attacks
can be a valid reason to deploy them instead of NNs in security critical
applications.

\bibliographystyle{unsrt}

\end{document}